Honghan Wu [1,2,3*], Karen Hodgson [4], Sue Dyson [4], Katherine I Morley [4,5,6], Zina M Ibrahim [4,7], Ehtesham Iqbal [4], Robert Stewart [4,5], Richard JB Dobson [4,7], and Cathie Sudlow [1,2]

[1] Centre for Medical Informatics, Usher Institute, University of Edinburgh, United Kingdom
[2] Health Data Research, University of Edinburgh, United Kingdom
[3] School of Computer and Software, Nanjing University of Information Science & Technology, China
[4] Institute of Psychiatry, Psychology & Neuroscience, King's College London, United Kingdom
[5] South London and Maudsley NHS Foundation Trust, London, UK
[6] Centre for Epidemiology and Biostatistics, Melbourne School of Global and Population Health, The University of Melbourne, Melbourne, Australia
[7] Health Data Research UK, University College London, United Kingdom

*Corresponding: 9 Little France Rd, Edinburgh EH16 4UX, United Kingdom; +44 (0)131 651 7882; honghan.wu@ed.ac.uk

# Efficiently Reusing Natural Language Processing Models for Phenotype-Mention Identification in Free-text Electronic Medical Records: Methodology Study

## Abstract

**Background:**
Many efforts have been put into the use of automated approaches, such as natural language processing (NLP), to mine or extract data from free-text medical records to construct comprehensive patient profiles for delivering better health-care. Reusing NLP models in new settings, however, remains cumbersome - requiring validation and/or retraining on new data iteratively to achieve convergent results.

**Objective:** The aim of this work is to minimize the effort involved in reusing NLP models on free-text medical records.

**Methods:** We formally define and analyse the model adaptation problem in phenotype-mention identification tasks. We identify "*duplicate waste*" and "*imbalance waste*", which collectively impede efficient model reuse. We propose a phenotype embedding based approach to minimize these sources of waste without the need for labelled data from new settings.

**Results:** We conduct experiments on data from a large mental health registry to reuse NLP models in four phenotype-mention identification tasks. The proposed approach can choose the best model for a new task, identifying up to 76% (duplicate waste), i.e. phenotype mentions without the need for validation and model retraining, and with very good performance (93-97% accuracy). It can also provide guidance for validating and retraining the selected model for novel language patterns in new tasks, saving around 80% (imbalance waste), i.e. the effort required in "*blind*" model-adaptation approaches.
**Conclusions:** Adapting pre-trained NLP models for new tasks can be more efficient and effective if the language pattern landscapes of old settings and new settings can be made explicit and comparable. Our experiments show that the phenotype-mention embedding approach is an effective way to model language patterns for phenotype-mention identification tasks and that its use can guide efficient NLP model reuse.

## Introduction

Compared to structured components of electronic health records (EHRs), free-text comprises a much deeper and larger volume of health data. For example, in a recent geriatric syndrome study[1], unstructured EHR data contributed a significant proportion of identified cases: 67.9% falls, 86.6% cases of visual impairment, and 99.8% cases of lack of social support. Similarly, in a study of co-morbidities using a database of anonymised EHRs of a psychiatric hospital in London (the South London and Maudsley NHS Foundation Trust (SLaM))[2], 1,899 cases of co-morbid depression and type 2 diabetes were identified from unstructured EHRs, while only 19 cases could be found using structured diagnosis tables. The value of unstructured records for selecting cohorts has also been widely reported[3,4]. Extracting clinical variables or identifying phenotypes from unstructured EHR data is, therefore, essential for addressing many clinical questions and research hypotheses [5–7].

Automated approaches are essential to surface such deep data from free-text clinical notes at scale. To make NLP tools accessible for clinical applications, various approaches have been proposed, including generic, user-friendly tools[8–10] and web services or cloud based solutions[11–13]. Among these approaches, perhaps the most efficient way to facilitate clinical NLP projects is to adapt pre-trained NLP models in new but similar settings[14], i.e. to re-use existing NLP solutions to answer new questions or to work on new data sources.. However, it is very often burdensome to reuse pre-trained NLP models. This is mainly because NLP models essentially abstract language patterns (i.e. language characteristics representable in computable form) and subsequently use them for prediction or classification tasks. These patterns are prone to change when the document set (corpus) or the text mining task (what to look up) changes. Unfortunately, when it comes to a new setting, it is uncertain which patterns have and have not changed. Therefore, in practice, random samples are drawn to validate the performance of an existing NLP model in a new setting and subsequently to plan the adaptation of the model based on the validation results.

Such "*blind*" adaptation is costly in the clinical domain because of barriers to data access and expensive clinical expertise needed for data labelling. The *"blindness"* to the similarities and differences of language pattern landscapes between the source (where the model was trained) and target (the new task) settings causes (at least) two types of potentially unnecessary, wasted effort, which may be avoidable. First, for those data in the target setting with the same patterns as in the source setting, any validation or retraining efforts are unnecessary because the model has already been trained and validated on these language patterns. We call this type of wasted effort the "***duplicate waste***". The second type of *waste* occurs if the distribution of new language patterns in the target setting is unbalanced, i.e., some - but not all - data instances belong to different language patterns. The model adaptation involves validating the model on these new data and further adjusting it when performance is not good enough. Without the knowledge of which data instances belong to which language patterns, data instances have to be randomly sampled for validation and adaptation. In most cases, a minimal number of instances of every pattern need to be processed so that convergent results can be obtained. This will usually be achieved via iterative validation and adaptation process, which will inevitably cause commonly used language patterns to be over represented, resulting in the model being over validated/retrained on such data. Such unnecessary effort on commonly used language patterns result from the pattern imbalance in the target setting, which unfortunately is the norm in almost all real world EHR datasets. We call this "***imbalance waste***".

The ability to make language patterns *visible* and comparable will address whether an NLP model can be adapted to a new task and, importantly, provide guidance on how to solve new problems effectively and efficiently through the *smart* adaptation of existing models. In this paper, we introduce a contextualised embedding model to *visualise* such patterns and provide guidance for reusing NLP models in phenotype-mention identification tasks. Here, a phenotype mention denotes an appearance of a word or phrase (representing a medical concept) in a document, which indicates a phenotype related to a person. We note two aspects of this definition:
1. Phenotype mention ≠ Medical concept mention. When a medical concept mentioned in a document does not indicate a phenotype relating to a person (e.g., cases in the last two rows of Table 1), it is not a phenotype mention.
2. Phenotype mention ≠ Phenotype. Phenotype (e.g. diseases and associated traits) is a specific patient characteristic[15], which is a patient level feature, e.g., a binary value indicating a patient is a smoker or not. However, for the same phenotype, a patient might have multiple phenotype mentions. For example, xxx is a smoker could be mentioned in different documents or even multiple times in one document, each of these appearances is a phenotype mention.

Table 1. The task of recognising contextualised phenotype mentions is to identify mentions of phenotypes from free-text records and also classify the context of each mention into 5 categories (listed in the 3rd column of Table 1). The last two rows

give examples of non-phenotype mentions - the two sentences are not describing incidents of a condition.

| Examples | Types of phenotype mentions | |
| --- | --- | --- |
| 49 year old man with *hepatitis c* | Contextualised mentions | positive mention |
| with no evidence of *cancer recurrence* | | negated mention |
| is concerning for local *lung cancer recurrence* | | hypothetical mention |
| PAST MEDICAL HISTORY: 1) *Atrial Fibrillation*, 2)... | | history mention |
| Mother was A positive, *hepatitis C carrier*, and ... | | mention of phenotype in another person |
| She visited the *HIV* clinic last week. | not a phenotype mention | |
| The patient asked for information about *stroke*. | not a phenotype mention | |

The focus of this work is to minimise the effort in reusing existing NLP model(s) in solving new tasks rather than proposing a novel NLP model for phenotype-mention identification. We aim to address the problem of NLP model transferability in the task of extracting mentions of phenotypes from free-text medical records. Specifically, the task is to identify above-defined phenotype mentions and the contexts in which they were mentioned[10] . (Table 1) explains and gives examples of contextualised phenotype mentions. The research question to be investigated is formally defined as follows.

Figure 1. Assess the transferability of a pre-trained model in solving a new task: discriminate between differently inaccurate mentions identified by the model in the new setting.

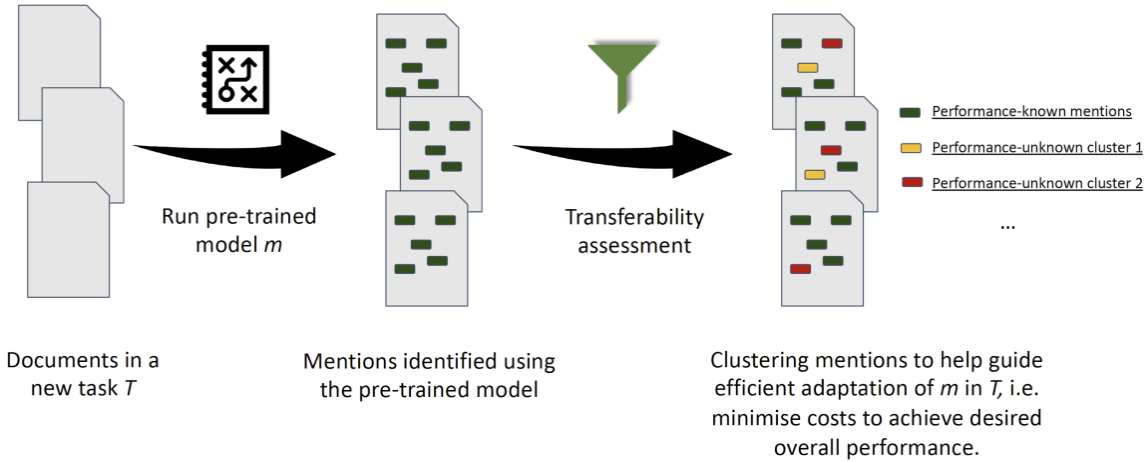

**Definition 1.** Given an NLP model (denoted as $m$) previously trained for some phenotype-mention identification task(s), and a new task (denoted as $T$, where either phenotypes to be identified are new or the dataset is new, or both are new), m is used in $T$ to identify a set of phenotype mentions - denoted as $S$. The research question (as illustrated in Figure 1) is how to partition $S$ to meet the following criteria:
1. a maximum *p-known* subset $S_{known}$ where $m$'s performance can be properly predicted using prior knowledge of $m$;
2. *p-unknown* subsets: $\{S_{u_1}, S_{u_2}, \ldots, S_{u_k}\}$, which meet the following criteria:
   a. $S_{u_1} \cup S_{u_2} \cup \ldots \cup S_{u_k} = S - S_{known}$;
   b. $\forall i, j \in [1..k], i \neq j, S_{u_i} \cap S_{u_j} = \emptyset$;
   c. $\forall i \in [1..k], S_{u_i}$ can be represented by a small number of instances $R_{u_i}$ so that $m$'s overall performance on $S_{u_i}$ can be predicted by its result on $R_{u_i}$;
   d. $k \ll |S| - |S_{known}|$.

The identification of '*p-known*' subset (**criterion 1**) will help eliminate "***duplicate waste***" by avoiding unnecessary validation and adaptation on those phenotype mentions. On the other hand, separating the rest of the annotations into '*p-unknown*' subsets allows processing mentions based on their *performance-relevant* characteristics separately, which in turn helps avoid "***imbalance waste***". The above **criterion 2.a** ensures completeness of coverage of all performance-unknown mentions, **2.b** ensures no overlaps between mention subsets so that no duplicated effort will be put on the same mentions. **Criterion 2.c** requires that the partitioning of the mentions is *performance-relevant*, meaning model performance on a small number of samples can be generalised to the whole subset that they are drawn from. Lastly, a small $k$ (**criterion 2.d**) enables efficient adaptation of a model.

## Methods

### Dataset & adaptable phenotype-mention identification models

Recently, we developed SemEHR[10] - a semantic search toolkit aiming to use interactive information retrieval functionalities to replace NLP building so that clinical researchers can use a browser based interface to access text mining results from a generic NLP model and (optionally) keep getting better results by iteratively feeding back to the system. A SLaM instance of this system has been trained for supporting 6 comorbidity studies (62,719 patients; 17,479,669 clinical notes in total), where different combinations of physical conditions and mental disorders are extracted and analysed. Multimedia Appendix 1 give details about the user interface and model performance. These studies effectively generated 23 phenotype-mention identification models and relevant labelled data (>7,000 annotated documents), which we use to study model transferability.

### Foundation of Proposed Approach

Our approach is based on the following assumption about a language pattern representation model.

**Assumption 1.** There exists a pattern representation model, A, for identifying language patterns of phenotype mentions with the following characteristics.
1. Each phenotype mention can be characterised by one and only one language pattern;
2. Patterns are largely shared by different mentions;
3. There is a deterministic association between NLP models' performances with such language patterns.

**Theorem 1.** Given $A$ - a pattern model meeting Assumption 1, $m$ - an NLP model, $T$ - a new task, let $P_m$ be the pattern set $A$ identifies from dataset(s) that $m$ was trained or validated on; let $P_T$ be the pattern set $A$ identifies from $S$ - the set of all mentions identified by $m$ in $T$. Then, the problem defined in Definition 1 can be solved by a solution, where $P_m \cap P_T$ is the '*p-known*' subset and $P_T - (P_m \cap P_T)$ is '*p-unknown*' subsets.

Proof of Theorem 1 can be found in the Multimedia Appendix 2. The rest of this section gives details of a realisation of $A$ using distributed representation models.

### Distributed representation for contextualised phenotype mentions

In computational linguistics, statistical language models are perhaps the most common approach to quantify word sequences, where a distribution is used to represent the probability of a sequence of words - $P(w_1, \ldots, w_n)$. Among such models, the bag-of-words (BOW) model[15] is perhaps the earliest and simplest, yet still widely-used and efficient in certain tasks[16]. To overcome BOW's limitations (e.g., ignoring semantic similarities between words), more complex models were introduced to represent word semantics [17–19]. Probably, the most popular alternative is the distributed representation model[20], which uses a vector space to model words so that word similarities can be represented as distances between their vectors. This concept has since been extensively followed up, extended and shown to significantly improve NLP tasks[21–26].

In original distributed representation models, the semantics of one word is encoded in one single vector, which makes it impossible to disambiguate different semantics or contexts that one word might be used in a corpus. Recently, various (bidirectional) Long Short Term Memory (LSTM) models were proposed to learn contextualised word vectors[27–29]. However, such linguistic contexts are not the phenotype contexts (see Table 1) that we seek in this paper.

Figure 2. The framework to learn contextualised phenotype embedding from labelled data that an NLP model m was trained or validated on.

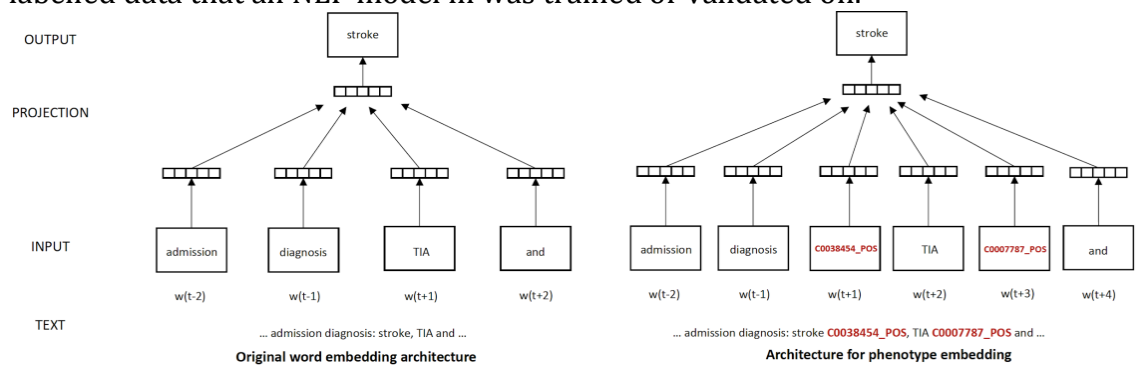

Inspired by the good properties of distributed representations for words, we propose a phenotype encoding approach that aims to model the language patterns of contextualised phenotype mentions. Compared to word semantics, phenotype semantics are represented in a larger context - at sentence or even paragraph level (e.g., *he worries about contracting* **HIV** - HIV here is a hypothetical phenotype mention). The key idea of our approach is to use explicit mark-ups to represent phenotype semantics in the text so that they can be learnt through an approach similar to word embedding learning framework.

(Figure 2) illustrates our framework for extending the continuous BOW word embedding architecture to capture the semantics of contextualised phenotype mentions. Explicit *mark-ups of phenotype mentions* are added to the architecture as placeholders for phenotype semantics. A mark-up (e.g., C0038454_POS) is composed of two parts: phenotype identification (e.g., C0038454) and contextual description (e.g., POS). The first part identifies a phenotype using a standardised vocabulary. In our implementation the Unified Medical Language System (UMLS)[30] was chosen for its broad concept coverage and the provision of comprehensive synonyms for concepts. The first benefit of using a standardised phenotype definition is that it helps in grouping together mentions of the same phenotype using different names. For example, using UMLS concept identification of C0038454 for STROKE helps combining together mentions using *Stroke, Cerebrovascular Accident, Brain Attack* and other 43 synonyms. The second benefit comes from the concept relations represented in the vocabulary hierarchy, which helps the transferability computation that we will elaborate later. The second part of a phenotype mention mark-up is to identify the mention context. Six types of contexts are supported: POS for *positive mention*, NEG for *negated mention*, HYP for *hypothetical mention*, HIS for *history mention*, OTH for *mention of the phenotype in another person* and NOT for *not a phenotype mention*.
There
The *phenotype mention mark-ups* can be populated using labelled data that NLP models were trained or validated on. In our implementation, the mark-ups were generated from the labelled subset of SLaM EHRs.

## Using phenotype embedding and their semantics for assessing model transferability

Figure 3. Architecture of phenotype embedding based approach for transferring pre-trained natural language processing models for identifying new phenotypes or application to new corpora. The word & phenotype embedding model is learnt from the training data of the re-usable models in its source domain (the task that m was trained for). No labelled data in the target domain (new setting) is required for the adaptation guidance.

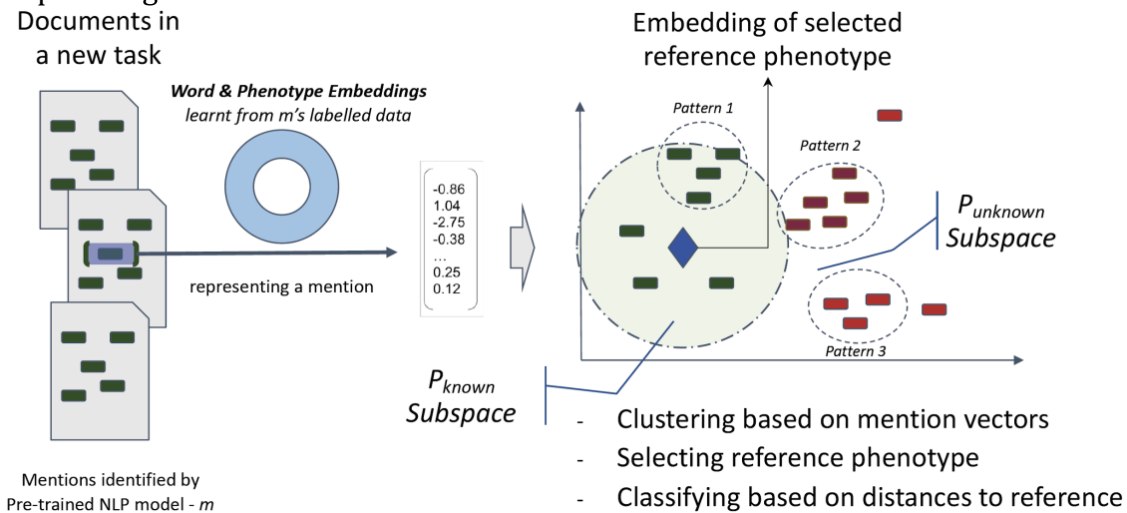

The embeddings learnt (including both word and contextualised phenotype vectors) are the building blocks underlying the language pattern representation model - $A$, as introduced at the beginning of this section, which is to compute $P_m$ (the landscape of language patterns that $m$ is familiar with) and $P_T$ (the landscape of language patterns in the new task $T$) for assessing and guiding NLP model adaptation for new tasks.

(Figure 3) illustrates the architecture of our approach. The double-circle shape denotes the embeddings learnt from the $m$'s labelled data. Essentially, the process is composed of two phases: (1) the documents from a new task (on the left of the figure) are annotated with phenotype mentions using a pre-trained model - $m$; (2) a classification task uses the above embeddings to assess each mention: whether it is an instance of *p-known* (something similar enough to what $m$ is familiar with) or any subset of *p-unknown* (something that is new to $m$). Specifically, the process is composed of the following steps.

1. **Vectorise phenotype mentions in a new task** Each mention in the new task will be represented as a vector of real numbers using the learnt embedding model to combine its surrounding words as context semantics. Formally,

   > let $s$ be a mention identified by $m$ in the new task, where $s$ can be represented by a function defined as follows:
   > $$v(s) = f(\vec{w}(t_{i-k}, \ldots, t_{i+k+l})) \quad (1)$$
   > where $\vec{w}$ is the embedding model to convert a word token into a vector, $t_j$ is the $j_{th}$ word in a document, $i$ is the offset of the first word of $s$ in the document, $l$ is the number of words in $s$ and $f$ is a function to combine a set of vectors into a result vector (we use *average* in our implementation).

   With such representations, all mentions are effectively put in a vector space (depicted as a two dimensional space on the right of the figure for illustration purposes).

2. **Identify clusters (language patterns) of mention vectors** In the vector space, clusters are naturally formed based on geometric distances between mention vectors. After trying different clustering algorithms and parameters, DBScan[31] was chosen on Euclidean distance in our implementation for vector clustering. Essentially, each cluster is a set of mentions considered to share the same (or similar enough) underlying language pattern, meaning language patterns in the new task are technically the vector clusters. We chose the cluster centroid (arithmetic mean) to represent a cluster (i.e., its underlying language pattern).

3. **Choose a reference vector for classifying language patterns** After clusters (language patterns) are identified, the next step is to classify them as *p-known* or subsets of *p-unknown*. We choose a reference vector-based approach - classifying using the distance to a selected vector. Such a reference vector is picked up (when the phenotype to be identified has been trained in $m$) or generated (when the phenotype is new to $m$) from the learnt phenotype embeddings the model $m$ has seen previously. Apparently, when the phenotype to be identified in the new task is new to $m$ (not in the set of phenotypes it was developed for), the reference phenotype needs to be carefully selected so that it can help to produce a sensible separation between *p-known* and *p-unknown* clusters. We use the semantic similarity (distance between two concepts in the UMLS tree structure) to choose the most similar phenotype from the phenotype list $m$ was trained for. Formally, the reference is chosen as follows.

> Let $c_p$ be the UMLS concept for a phenotype to be identified in the new task and $C_m$ be the set of phenotype concepts that m was trained for, the reference phenotype choosing function is:
> $$R(c_p, C_m) = argmin_{c \in C_m} D(c, c_p) \quad (2)$$
> where $D$ is a distance function to calculate the steps between two nodes in the UMLS concept tree.

Once the reference phenotype has been chosen, the reference vector can be selected or generated (e.g., use the average) from this phenotype's contextual embeddings.

4. **Classify language patterns to guide model adaptation** Once the reference vector has been selected, clusters can be classified based on the distances between their centroids (representative vectors of clusters) and the reference vector. Once a distance threshold is chosen, this distance-based classification partitions the vector space into two subspaces using the reference vector as the centre: the sub-space whose distance to the centre is less than the threshold is called *p-known* sub-space and the remainder is *p-unknown* sub-space. The union of clusters whose centroids are within *p-known* sub-space is *p-known* meaning $m$'s performances on them can be predicted without further validation (**removing *duplicate waste***). Other clusters are *p-unknown* clusters. $m$ can be validated and/or further trained on each *p-unknown* cluster separately instead of blindly across all clusters. This will **remove *imbalance waste***.

## Results

**Associations between embedding based language patterns and model performances**
As stated in the beginning of Method section, our approach is based on 3 assumptions about language patterns. Therefore, it is essential to quantify to what extent the language patterns identified by our embedding based approach meet these assumptions. The first assumption - a phenotype mention can be assigned to one and only to one language pattern - is met in our approach, since (a) (Equation 1) is a One-to-One function; (b) DBScan algorithm (the vector clustering function chosen in our implementation) is also a One-to-One function. Assumption 2 can be quantified by the percentage of mentions that can be assigned to a cluster. This percentage can be increased by increasing the Epsilon(EPS) parameter (the maximum distance between two data items for them to be considered as in the same neighbourhood) in DBScan. However, the degree to which mentions are clustered together needs to be balanced against the consequence of reduced ability to identify performance-related language patterns, which is the third assumption - associations between language patterns and model performance. To quantify such an association, we propose a metric called Bad Guy Separate Power (SP for short), as defined in (Equation 3) below. The aim is to measure to what extent a clustering can assemble incorrect data items (false positive mentions of phenotypes) together.

Let C be a set of binary data items - $\forall c_i \in C, T(c_i) \in \{t, f\}$ ($t$ – stands for true; $f$ – stands for false), given a clustering result $\{C_1, \ldots, C_k | C_1 \cup C_2 \ldots \cup C_k = C\}$, its separate power for $f$ typed data items is defined as follows.

$$SP(\{C_1, \ldots, C_k\}, f) = \frac{\sum_{i=1}^{k} \frac{|\{c_i | c_i \in C, T(c_i) = f\}|^2}{|C_i|}}{|\{c_i | c_i \in C, T(c_i) = f\}|} \quad (3)$$

In our scenario, we would like to see clustering being able to separate easy cases (where good performance is achieved) from difficult cases (where performance is poor) for a model $m$.

Figure 4. Clustered Percentage vs Separate Power on difficult cases. The X axis is the EPS parameter of the DBScan clustering algorithm - the longest distance between any two items within a cluster; the Y axis is the percentage. Two types of changing information (as functions of EPS) are plotted on each sub-figure: clustered percentage (solid line) and Separate Power (SP) on incorrect cases (false positive mentions of phenotypes). The latter has two series: (1) SP by chance (dash dotted line) - when clustering by randomly selecting mentions; (2) SP by clustering using phenotype embedding (dashed line). $N$ - number of all mentions; $N_f$ – number of false positive mentions.

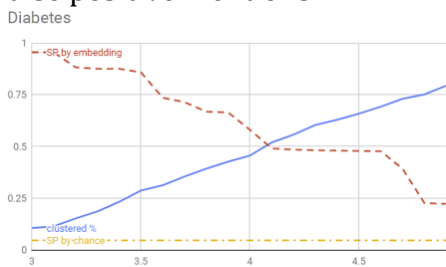

(a) Diabetes (C0011849): $N = 268, N_f = 23$

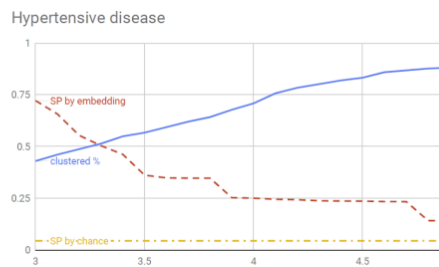

(b) Hypertensive disease (C0020538): $N = 238, N_f = 13$

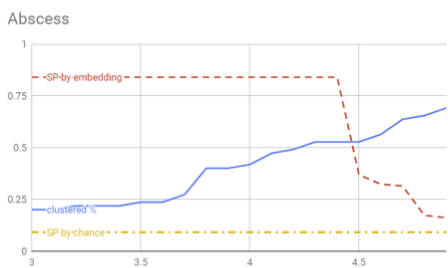

(c) Abscess (C0000833): $N = 86, N_f = 11$

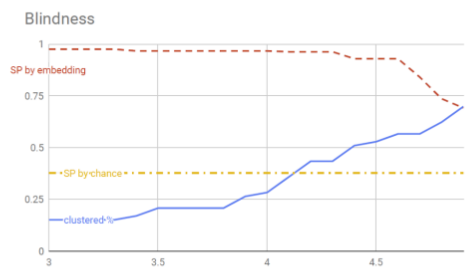

(d) Blindness (C0456909): $N = 58, N_f = 27$

To quantify the clustering percentage, the ability to separate mentions based on model performances and the interplay between the two, we conducted experiments on selected phenotypes by continuously increasing the clustering parameters - EPS from a low level. (Figure 4) shows the results. In this experiment, we label mentions into two types: correct and incorrect using SemEHR labelled data on the SLaM corpus. Specifically, for the mention types in (Table 1), incorrect mentions are those denoted 'not-a-phenotype-mention' and the remainder are labelled as correct.

We chose incorrect as the $f$ in (equation 3) meaning that we evaluate the separate power on incorrect mentions. Four phenotypes were selected for this evaluation: *Diabetes* and *Hypertensive disease* were selected because they were most validated phenotypes; *Abscess* (with 13% incorrect mentions) and *Blindness* (with 47% incorrect mentions) were chosen to represent NLP models with different levels of performance. The figure shows a clear trend in all cases that as EPS increases the clustered percentage increases but with decreasing separate power. This confirms a trade-off between the coverage of identified language patterns and how good they are. Regarding separate power, the performance on two selected common phenotypes (Figure 4a and 4b) is generally worse than for the other phenotypes - starting with lower power, which decreases faster as EPS increases. The main reason is that the difficult cases (mentions with poor performance) in the two commonly encountered phenotypes are relatively rare (Diabetes: 8.5%; Hypertensive disease: 5.5%). In such situations, difficult cases are harder to separate because their patterns are under-represented. However, in general, compared to random clustering, the embedding based clustering approach brings in much better separate power in all cases. This confirms a high level association between identified clusters and model performance. In particular, when the proportion of difficult cases reaches near 50% (Figure 4d), the approach can keep *SP* values almost constantly near 1.0 when EPS increases. This means it can almost always group difficult cases in their own clusters.

### Model adaptation guidance evaluation

Technically, the guidance to model adaptation is composed of two parts: avoid **duplicate waste** (skip validation/training efforts on cases the model is already familiar with); and avoid **imbalance waste** (group new language patterns together so that validation/continuous training on each group separately can be more efficient than doing it over the whole corpus). To quantify the guidance effectiveness, the following metrics are introduced.

- **Duplicate Waste.** This is the number of mentions whose patterns fall into what the model m is familiar with. The quantity $\frac{|\{s|pattern(s) \in P_m \cap P_T\}|}{|S|}$ is the proportion of mentions which needs no validation or retraining before reusing $m$.
- **Imbalance Waste.** To achieve convergence performance, an NLP model needs to be trained on a minimal number (denoted as $e$) of samples from each language pattern. Calling the language pattern set in a new task as $C = \{C_1, \ldots, C_k\}$, the following equation counts the minimum number of samples needed to achieve convergent results in *'blind'* adaptations.

$$Conv\_Sampling(C, e) = max_{i=1}^{k} \frac{|S|}{|C_i|} \times min(|C_i|, e) \qquad (4)$$

When the language patterns are identifiable, the **Imbalance Waste** that can be avoided is quantified as $Conv\_Sampling(C, e) - \sum_{i=1}^{k} min(|C_i|, e)$.
- **Accuracy.** To evaluate whether our approach can really identify familiar patterns, we quantify the accuracy of those within-threshold clusters and also those within-threshold single mentions that are not clustered. Both macro-accuracy (average of all cluster accuracies) and micro-accuracy (overall accuracy) are used - detailed explanations at [32].

([Figure 5](#)) shows the results of our NLP model adaptation guidance on 4 phenotype identification tasks. For each new phenotype identification task, the NLP model (pre-)trained for the semantically most similar (defined in Equation 2) phenotype was chosen as the reuse model. Models and labelled data for the four pairs of phenotypes were selected from six physical comorbidity studies on SLaM data. Figure 5 shows that identified mentions have a high proportion of avoidable duplicate waste in all 4 cases: diabetes and heart attack start with 50%; stroke and multiple sclerosis > 70%. Such avoidable duplicate waste decreases when the threshold increases. The threshold is on similarity instead of distance, meaning that new patterns need to be more similar to the reuse model's embeddings to be counted as familiar patterns. Therefore, it is understandable that duplicate waste decreases in such scenarios. In terms of accuracy, one would expect this to increase as only more similar patterns are left when threshold increases. However, interestingly, in all cases, both macro and micro-accuracies decrease slightly before increasing to reach near 1.0. This is a phenomenon worth future investigation. In general, the changes in accuracy are small (.03 to .08), while accuracy remains high (>.92). Given these observations, the threshold is normally set at .01, to optimise the avoidance of duplicate waste with minimal effect on accuracy. Specifically, in all cases, more than half of the identified mentions (50%+ for subfigure 5a and 5b; 70%+ for 5c and 5d) do not need any validation/training to obtain accuracy of >0.95. In terms of effective adaptation on new patterns, the percentages of avoidable imbalance waste in all cases are around 80% confirming that a much more efficient retraining on data can be achieved through language pattern-based guidance.

Figure 5. Identifying new phenotypes by reusing NLP models pre-trained for semantically-close phenotypes: the four pairs of phenotype-mention identification models are chosen from SemEHR models trained on SLaM data; DBScan EPS value: 3.8; Imbalance Waste is calculated on e = 3 meaning at least 3 samples are needed for training from each language pattern. The X axis is the similarity threshold, ranging from 0.0 to 0.8; the Y axes, from top to bottom, are: the proportion of duplicate waste saved over total number of mentions; macro-accuracy; micro-accuracy.

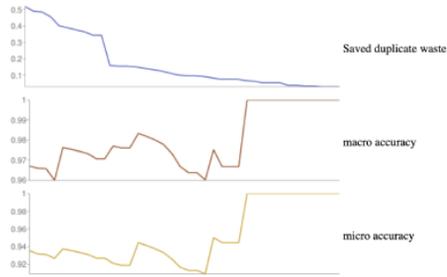

**(a)** New task: *Diabetes (C0011849)*;
Reuse model: *Type 2 Diabetes(C0011860)*;
#Mentions/#not-a-mention: 268/23;
#Cluster:15;
Saved Imbalance Waste: 40 or 83%

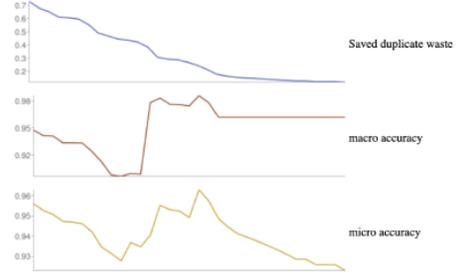

**(b)** New task:*Stroke(C0038454)*;
Reuse model: *Heart Attack(C0027051)*;
#Mentions/#not-a-mention: 238/13;
#Cluster:16;
Saved Imbalance Waste: 39 or 82%

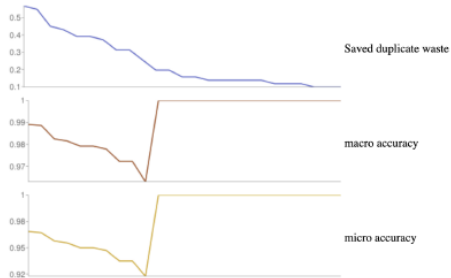

**(c)** New task: *Heart Attack(C0027051)*;
Reuse model: *Infarct(C0021308)*;
#Mentions/#not-a-mention: 54/11;
#Cluster:5;
Saved Imbalance Waste: 11 or 78%

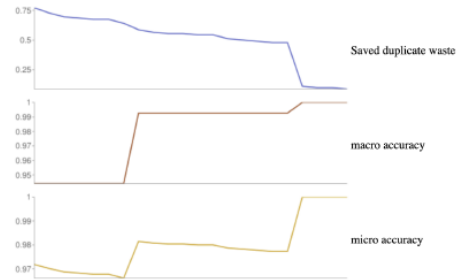

**(d)** New task:*Multiple Sclerosis(C0026769)*;
Reuse model: *Myasthenia Gravis(C0026896)*;
#Mentions/#not-a-mention: 104/4;
#Cluster:5;
Saved Imbalance Waste: 14 or 85%

### Effectiveness of phenotype semantics in model reuse

When considering NLP model reuse for a new task, if there is no existing model that has been developed for the same phenotype-mention identification task, our approach will choose a model trained for a phenotype that is most semantically similar to it (based on Equation 2). To evaluate the effectiveness of such semantic relationships in reusing NLP models, we conducted experiments on the previous four phenotypes by using phenotype models with different levels of semantic similarities. (Table 2) shows the results. In all cases, reusing models trained for more similar phenotypes can identify more ***duplicate waste*** using the same parameter settings. The first three cases in the table can also achieve better accuracies, while *multiple sclerosis* had slightly better accuracy by reusing the *diabetes* model than the more semantically-similar *myasthenia gravis*. However, the latter identified 46% more ***duplicate waste***.

Table 2. Comparisons of the performance of reusing models with different semantic similarity levels. More similar ones are marked with *. Similarity threshold: 0.01; DBScan EPS: 0.38. Reusing models trained for more (semantically) similar phenotypes achieved adaptation results that with less effort (more duplicate waste identified) in all cases and were also more accurate in 3 out of 4 cases.

| Model reuse cases | duplicate waste | macro-accuracy | micro-accuracy |
|---|---|---|---|
| *Diabetes* by *Type 2 Diabetes*\* | **0.502** | **0.966** | **0.933** |
| *Diabetes* by *Hypercholesterolaemia* | 0.477 | 0.965 | 0.930 |
| *Stroke* by *Heart attack*\* | **0.711** | **0.948** | **0.955** |
| *Stroke* by *Fatigue* | 0.220 | 0.884 | 0.938 |
| *Heart attack* by *Infarct*\* | **0.569** | **0.989** | **0.966** |
| *Heart attack* by *Bruise* | 0.529 | 0.821 | 0.889 |
| *Multiple Sclerosis* by *Myasthenia Gravis*\* | **0.761** | 0.944 | 0.971 |
| *Multiple Sclerosis* by *Diabetes* | 0.522 | **0.993** | **0.979** |

## Discussion

### Principal Results

Automated extraction methods (as surveyed recently by Ford and et. al.[33]), many of which are made freely available and open source, have been intensively investigated in mining free-text medical records[10,34–36]. To provide guidance in the efficient reuse of pre-trained NLP models, we have here proposed an approach that can automatically (i) identify easy cases in a new task for the reused model, on which it can achieve good performance with high confidence; (ii) classify the remainder of the cases so that the validation or retraining on them can be conducted much more efficiently, compared to adapting the model on all cases. Specifically, in four phenotype-mention identification tasks, we have shown that 50-79% of all mentions are identifiably easy cases, for which our approach can choose the best model to reuse, achieving more than 93% accuracy. Furthermore, for those cases that need validation or retraining, our approach can provide guidance that can save 78-85% of the validation/retraining effort. A distinct feature of this approach is that it requires no labelled data from new settings, which enables very efficient model adaptation - as shown in our evaluation: zero effort to obtain >93% accuracy among the majority (>63% in average) of the results.

### Limitations

In this study, we did not evaluate the recall of adapted NLP models in new tasks. Although the models we chose can generally achieve very good recall for identifying physical conditions (96-98%) within the SLaM records, investigating the transferability on recalls is an important aspect of NLP model adaptation.

The model reuse experiments were conducted on identifying new phenotypes on document sets that had not previously been seen by the NLP model. However, these documents were still part of the same (SLaM) EHR system. To fully test the generalisability of our approach will require evaluation of model reuse in a different EHR system, which will require a new set of access approvals as well as information governance approval for the sharing of embedding models between different hospitals.

We chose a phenotype embedding model to represent language patterns. One reason is that we have limited number of manually annotated data items. The word embedding approach is unsupervised and the word-level "semantics" learnt from the whole corpus can help group similar words together in the vector space so that can help improve the phenotype level clustering performances. However, thorough comparisons between different language pattern models are needed to reveal whether other approaches, in particular simpler or less-computing-intensive approaches can achieve similar or different performances.

In addition, implementation-wise, vector clustering is an important aspect of this approach. We have compared DBScan with k-nearest neighbors algorithm in our model, which revealed that DBScan could achieve better SP powers in most scenarios. Using a 64-bit Windows 10 server with 16G memory and 8 core CPUs (3.6 GHz), DBScan uses 200M memory and takes 0.038 seconds on near 300 data points on average of 100 executions. However, it is worth in-depth comparisons between more clustering algorithms. In particular, a larger datasets might be needed to compare the clustering performances on both computational aspect and SP powers.

**Comparison with Prior Work**

NLP model adaptation aims to adapt NLP models from a source domain (with abundant labelled data) to a target domain (with limited labelled data). This challenge has been extensively studied in the NLP community[37–41]. However, most existing approaches assume a single language model (e.g. a probability distribution) from each domain. This limits the ability to identify and subsequently deal differently with data items with different language patterns. Such a limitation prevents fine-grained adaptations, such as the reuse or adaptation of one NLP model on those items for which it performs well, and the re-training of the same model or reuse of other models on those items for which the original NLP model performs poorly. By contrast, our work aimed to depict the language patterns (i.e. different language models) of both source and target domains, and subsequently provide actionable guidance on reusing models based on these fine-grained language patterns. Further, very few NLP model reuse studies have focused on free-text in electronic medical records. Indeed, to the best of our knowledge, this work is among the first to focus on model reuse for phenotype-mention identification tasks on real-world free-text electronic medical records.

Modelling language patterns has been investigated for different applications, such as the k-Signature approach[42] for identifying unique "signatures" of micro-message authors. This paper models language patterns for characterising "landscape" of phenotype mentions. One main difference is that we do not know how many clusters (or "signatures") of language patterns exist in our scenario. Technically, we use phenotype embeddings to model such patterns and, particularly, utilise phenotype semantic similarities (based on ontology hierarchies) for reusing learnt embeddings when necessary.

## Conclusions

Making fine-grained language patterns visible and comparable (in computable form) is the key to support 'smart' NLP model adaptation. We have shown that the phenotype embedding based approach proposed in this paper is an effective way to achieve this. However, our approach is just one way to model such fine-grained patterns. Investigating novel pattern representation models is an exciting research direction to enable automated NLP model adaptation and composition (i.e. combining various models together) for efficiently mining free-text electronic medical records in new settings with maximum efficiency and minimal effort.

## Acknowledgements
This research was funded by Medical Research Council / Health Data Research UK Grant (MR/S004149/1); Industrial Strategy Challenge Grant (MC_PC_18029); the National Institute for Health Research (NIHR) Biomedical Research Centre at South London and Maudsley NHS Foundation Trust and King's College London. The views expressed are those of the authors and not necessarily those of the NHS, the NIHR or the Department of Health and Social Care.

## Ethical approval and informed consent
De-identified patient records were accessed through the Clinical Record Interactive Search (CRIS) at the Maudsley NIHR Biomedical Research Centre, South London and Maudsley (SLaM) NHS Foundation Trust. This is a widely used clinical database with a robust data governance structure which has received ethical approval for secondary analysis (Oxford REC 18/SC/0372).

## Conflicts of Interest
none declared

## Abbreviations
EHR: Electronic Health Record
LSTM: Long Short Term Memory
NLP: Natural Language Processing
SemEHR: a semantic search toolkit that can be trained for different studies
SLaM: South London and Maudsley NHS Foundation Trust

## Data availability statement
The clinical notes are not sharable in the public domain. However, interested researchers can apply for research access through

https://www.maudsleybrc.nihr.ac.uk/facilities/clinical-record-interactive-search-cris/. The natural language processing tool, models and code of this work are available at https://github.com/CogStack/CogStack-SemEHR.

**Multimedia Appendix 1 User interface and model performances of phenotype NLP models**

**Multimedia Appendix 2 Proof of Theorem 1**